\documentclass[conference, final, 10pt]{IEEEtran}
\IEEEoverridecommandlockouts
\usepackage{cite}

\usepackage{amsmath,amssymb,amsfonts}
\usepackage{algorithmic}
\usepackage{graphicx}
\usepackage{textcomp}
\usepackage{xcolor}
\usepackage{cleveref}
\usepackage{amsmath,amsfonts,amssymb,amsthm}
\usepackage{textcomp}
\usepackage{nicefrac}       
\usepackage{placeins}
\usepackage{tikz}
\usetikzlibrary{bayesnet}
\newcommand{\bff}{\mathbf}

\newcommand{\bmu}{\boldsymbol{\mu}}
\newcommand{\bsigma}{\boldsymbol{\sigma}}
\newcommand{\bpi}{\boldsymbol{\pi}}
\newcommand{\RN}[1]{%
  \textup{\uppercase\expandafter{\romannumeral#1}}%
}
\DeclareMathOperator{\EX}{\mathbb{E}}
\def\BibTeX{{\rm B\kern-.05em{\sc i\kern-.025em b}\kern-.08em
    T\kern-.1667em\lower.7ex\hbox{E}\kern-.125emX}}
\begin{document}

\title{Disentangling and Learning Robust Representations with Natural Clustering\\
}

\author{\IEEEauthorblockN{Javier Antor\'{a}n}
\IEEEauthorblockA{\textit{ViVoLab, Arag\'{o}n Institute for Engineering Research (I3A)} \\
\textit{University of Zaragoza}\\
Zaragoza, Spain \\
javier.a.es@ieee.org}
\and
\IEEEauthorblockN{Antonio Miguel}
\IEEEauthorblockA{\textit{ViVoLab, Arag\'{o}n Institute for Engineering Research (I3A)}\\
\textit{University of Zaragoza}\\
Zaragoza, Spain\\
amiguel@unizar.es}
}

\maketitle

\begin{abstract}
Learning representations that disentangle the underlying factors of variability in data is an intuitive way to achieve generalization in deep models. In this work, we address the scenario where generative factors present a multimodal distribution due to the existence of class distinction in the data. We propose N-VAE, a model which is capable of separating factors of variation which are exclusive to certain classes from factors that are shared among classes. This model implements an explicitly compositional latent variable structure by defining a class-conditioned latent space and a shared latent space. We show its usefulness for detecting and disentangling class-dependent generative factors as well as its capacity to generate artificial samples which contain characteristics unseen in the training data.
\end{abstract}


\section{Introduction}
Disentangled representations, defined as those where each latent variable is responsible for only one generative factor in the data while being relatively invariant to other factors \cite{Bengio_representation_learning}, allow for more easily interpretable models, \cite{higgins_scan}, and facilitate generalization to previously unseen combinations of features, \cite{Hierarchical_Disentangled}. They also generalize to new domains that share underlying factors of variation, \cite{irina_darla}.

Typically, variational autoencoder (VAE) \cite{VAE} based disentanglement methods impose an isotropic Gaussian prior on their latent space. However, any dataset that can be separated into classes will present some form of multimodality. Despite the unimodal prior, in these scenarios, the inferred posterior is often multimodal. This creates dependencies between latent dimensions and reduces reconstruction accuracy, \cite{vamp}. 

\begin{figure}[t]
\vskip 0.05in
\begin{center}
\centerline{\includegraphics[width=1.05\columnwidth]{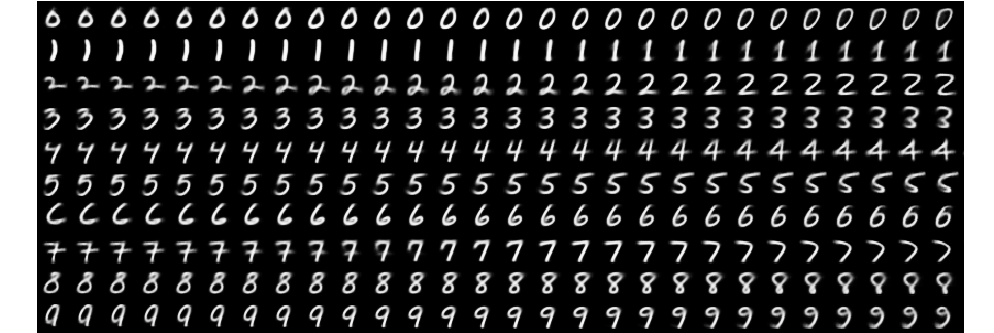}}
\caption{MNIST class-dependent latent traversals produced by N-VAE. Generative factors which are exclusive to each mode of the input data distribution are disentangled from factors which are shared across classes.}
\label{fig:class_dependent_MNIST}
\end{center}
\vskip -0.1in
\end{figure}

\begin{figure*}[ht]
\begin{center}
\centerline{\includegraphics[width=0.87\linewidth]{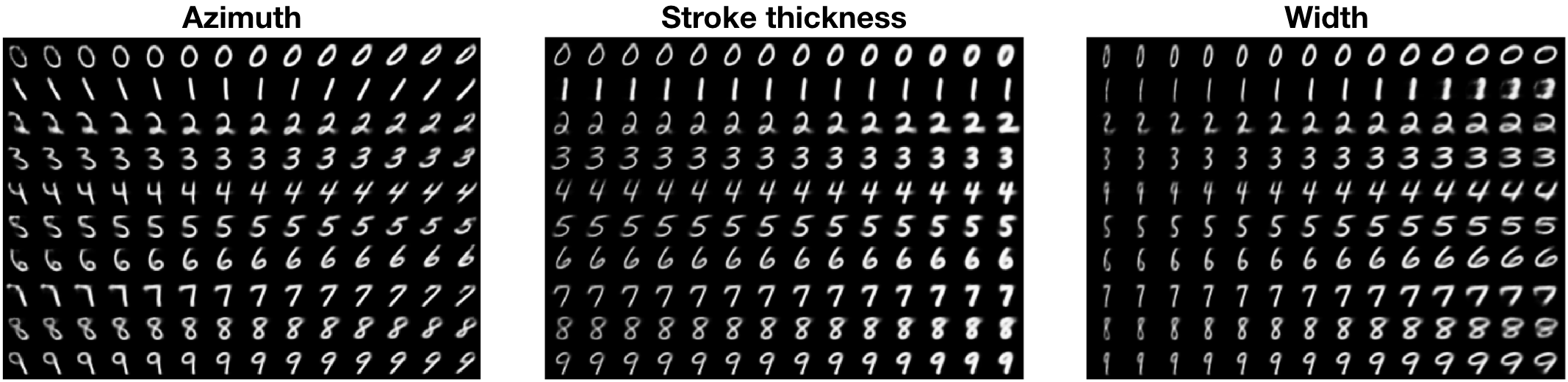}}
\caption{Traversals of N-VAE's shared latent space in the range $(-3, 3)$ when trained on the MNIST dataset.}
\label{fig:MNIST_shared_latents}
\end{center}
\vskip -0.2in
\end{figure*}

Moreover, there can exist underlying factors of variation in data that are exclusive to certain classes. An example of this phenomenon is found in the MNIST dataset. The number seven can be drawn with a horizontal line crossing its stem or without it. The number two can have a flat base or one with a loop. We show how these class-exclusive factors are disentangled by our proposed model in \cref{fig:class_dependent_MNIST}. Other factors of variability, such as stroke thickness or azimuth, are shared among all classes, as can be seen in \cref{fig:MNIST_shared_latents}.

The natural clustering prior \cite{Bengio_representation_learning} states that probability mass concentrates on a set of low dimensional manifolds. Each manifold is associated with a possible outcome of categorically distributed phenomena in the data, such as class membership. Our model, natural clustering VAE (N-VAE), conditions some of its latent variables on samples from a categorical distribution. By incorporating inductive biases about compositional structure in the data, it is able to capture class-dependent factors of variation which are usually ignored by other generative modelling techniques. 

Our main contributions are the following: 1) We formulate a lower bound on the joint log-likelihood of inputs and class labels. It can be optimized approximately by simply adding a classification objective to the standard VAE cost function. 2) We propose N-VAE, a model that is capable of capturing both generative factors which are class-dependent and ones that are shared among classes. To the best of our knowledge, we are the first to disentangle these two types of factors. 3) We show how N-VAE can be used for detection and disentanglement of mode-dependent factors of variation. We also evaluate the expressivity\footnote{We understand expressivity as the number of possible input configurations that are explained by a model, \cite{Bengio_representation_learning}.} of N-VAE relative to other class-conditioned generative models by comparing the usefulness of their generated samples for dataset augmentation.

\section{Preliminaries}
We first give a brief overview of the variational autoencoder, as N-VAE builds upon this model. We then give a recap of some recent techniques for disentangling in VAEs. We show that, despite the popularity of these models, they are not well equipped to disentangle multimodally distributed generative factors. 
\subsection{The Variational Autoencoder}
The VAE approximates the intractable posterior over a set of latent variables $p(\bff{z}|\bff{x})$ with an approximate distribution $q(\bff{z}|\bff{x})$. $D_{KL}( q(\bff{z}|\bff{x})\,\|\,p(\bff{z}|\bff{x}) )$ is minimized by optimizing the Evidence Lower Bound (ELBO) $\mathcal{L}(\bff{x}) \leq \log{p(\bff{x})}$:
\begin{align}\label{eq:ELBO}
    \mathcal{L}(\bff{x}) = - \underbrace{D_{KL} ( q(\bff{z}|\bff{x})\,\|\,p(\bff{z}) )}_{\text{Regularization cost}} + \underbrace{\EX_{q(\bff{z}|\bff{x})}[\log p(\bff{x}|\bff{z})]}_{\text{Reconstruction cost}}
\end{align}
$\mathcal{L}(\bff{x})$ becomes equal to $\log{p(\bff{x})}$  when $q(\bff{z}|\bff{x})\!=\!p(\bff{z}|\bff{x})$. The VAE framework parametrises both $p(\bff{x}|\bff{z})$ and $q(\bff{z}|\bff{x})$ with neural networks. Optimization is performed through gradient descent, using the \textit{reparameterisation trick}, \cite{VAE, otro_vae}.

The encoder, $q_{\phi}(\bff{z}|\bff{x})$, is an approximate inference model which parametrises a distribution over the latent variables $\bff{z}$. A Gaussian with diagonal covariance is usually selected: $\mathcal{N}(\bff{z}; \bmu(\bff{x};\phi), \bsigma(\bff{x};\phi)\!\cdot\!\bff{I})$. The decoder, $p_{\theta}(\bff{x}|\bff{z})$, is a generative model that reconstructs the input $\bff{x}$ from the latent variables $\bff{z}$. The prior over the latent variables is usually chosen to be $p(\bff{z})\!=\!\mathcal{N}(\bff{z};\,\bff{0}, \bff{I})$.
\subsection{Regularization-based approaches to disentangled VAEs}

The authors of \cite{BVAE} and \cite{understanding_BVAE} propose $\beta$-VAE. This model reweights the ELBO by multiplying the regularization cost by $\beta>1$. This pushes the approximate posterior to be closer to the factorial prior. In \cite{Disentangling_by_Factorising, Auto-Encoding_Total_Correlation, Isolating_Sources_of_Disentanglement}, the ELBO's regularization cost is decomposed into the mutual information between inputs and latent variables and the total correlation of the latent space, \cite{OG_total_correlation}. The authors propose different modifications to the regularization term of the VAE objective, all of which aim to minimize the total correlation without decreasing the mutual information. In \cite{ibm_disentanglement}, an approach that penalizes the L2 distance between the second order moments of the aggregate approximate posterior and factorial prior is proposed.

\begin{figure}[htb]
\begin{center}
\centerline{\includegraphics[width=1\columnwidth]{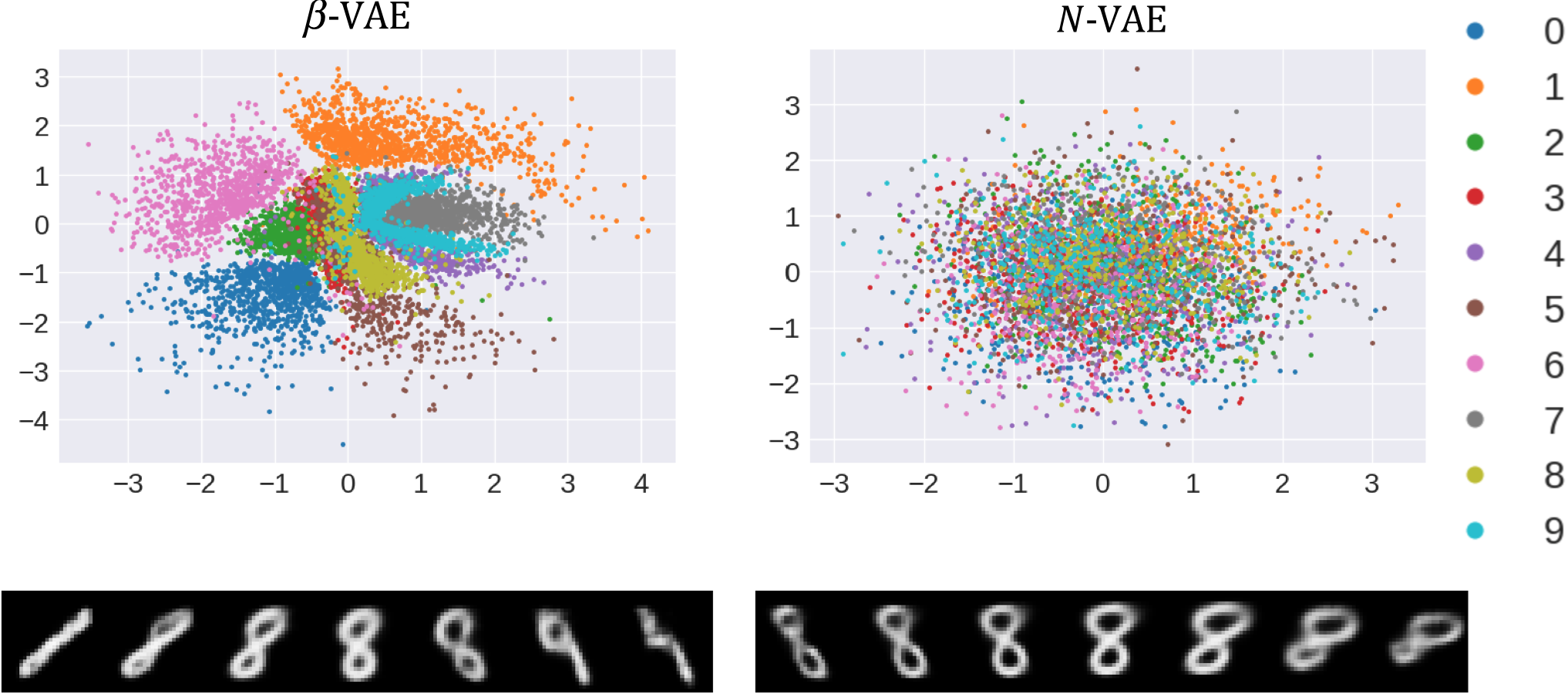}}
\caption{Latent variables inferred from MNIST's test set with $\beta$-VAE ($\beta\!=\!5$) and N-VAE. For N-VAE, the plot shows the shared latent variables $\bff{z}_{s}$. Latent spaces are chosen to be two-dimensional for ease of viewing. Below each distribution, we show latent traversals for both model's azimuth latent variable.}
\label{fig:beta_and_my_latent_space}
\end{center}
\vskip -0.1in
\end{figure}

\subsection{The problem with disentangling multimodal factors}\label{sec:why_dissentangling_is_hard}
In \cref{fig:beta_and_my_latent_space}, we show that the distribution of latent variables inferred by $\beta$-VAE \cite{BVAE} on the MNIST test set is multimodal. Roughly, one mode belongs to each digit. When traversing any given dimension of the latent space, we are likely to pass through low probability regions, generating poor quality images, and eventually switch digits.

In models with unimodal priors, such as the ones proposed in \cite{BVAE, understanding_BVAE, Disentangling_by_Factorising, Auto-Encoding_Total_Correlation, Isolating_Sources_of_Disentanglement, ibm_disentanglement}, multimodally distributed explanatory factors tend to occupy multiple dimensions of the latent space, becoming entangled with other attributes. We posit that expressing mode-dependent generative factors, such as the ones shown in \cref{fig:class_dependent_MNIST}, can result in a multimodal approximate posterior and thus, a high cost under the ELBO's KL term. These factors are often only learned if their impact on the reconstruction accuracy is significant enough to mitigate the large KL cost.

\section{Proposed Method} \label{sec:model}
Let $\{\bff{x}^{(i)}, y^{(i)}\}^{N}_{i=1}$ be a dataset consisting of $N$ \textit{i.i.d} observations of some variable $\bff{x}^{(i)}\,{\in}\,\mathbb{R}^{D}$ and its class label be $y^{(i)} \in \{1,\,... ,\,L\}$. We propose a generative model which describes $\bff{x}^{(i)}$ as being generated by a set of latent variables $\bff{z}^{(i)}$ and a class label $y^{(i)}$. The class probabilities are given by $\bpi^{(i)}$. Unlike \cite{kingma-semisupervised}, we treat $\bpi^{(i)}$ as a continuous latent variable over which we perform inference. We will omit the superscript $^{(i)}$ in order to maintain notation uncluttered in the rest of this paper. The joint distribution of our variables can be factorized as: $p(\bff{x}, y, \bff{z}, \bpi) = p(\bff{x}| y, \bff{z}) p(y | \bpi) p(\bpi) p(\bff{z})$.

\subsection{A Lower Bound on the Joint Log-Likelihood}
We formulate a variational lower bound on $\log p(\bff{x}, y)$:
\begin{align}
\log p(\bff{x}, y) \geq \mathcal{L}(\bff{x}, y) &= \EX_{q(\bff{z}, \bpi|\bff{x}, y)} [-\log q(\bff{z}, \bpi|\bff{x}, y)\notag\\
&+ \log p(\bff{x}, y, \bff{z}, \bpi)]\label{eq:jointELBO}
\end{align}
This expression can be expanded and rewritten as \cref{eq:ELBOdecomposed}. See appendix A. for a complete derivation.
\begin{gather}
\mathcal{L}(\bff{x}, y) =\EX_{q(\bff{z}|\bff{x})}[\log p(\bff{x}| y, \bff{z})] - D_{KL}(q(\bff{z} |\bff{x})\,\|\,\ p(\bff{z}))\notag\\
- D_{KL}(q(\bpi|\bff{x})\,\|\, p(\bpi|y)) + \log p(y)\label{eq:ELBOdecomposed}
\end{gather}
$\EX_{q(\bff{z}|\bff{x})}[\log p(\bff{x}| y, \bff{z})]$ is a reconstruction objective for our data vector $\bff{x}$ given $\bff{z}$ and $y$. $D_{KL}(q(\bpi|\bff{x})\,\|\, p(\bpi|y))$ penalizes the approximate distribution of $\bpi$ given $\bff{x}$ for differing from the true posterior of $\bpi$ given the class label $y$. This item can be interpreted as a classification objective. Its value will go to $0$ as $q(\bpi|\bff{x})$ approaches a one-hot vector with $\pi_{y}\!=\!1$. It is the most important difference between our lower bound and that of a standard VAE. $\log p(y)$ is constant with respect to our model's parameters.

\subsection{Proposed Model: N-VAE}
We place a zero mean isotropic unit variance Gaussian prior over $\bff{z}$: $p(\bff{z}) = \mathcal{N(\bff{z};\,\bff{0}, \bff{I})}$. We model the conditional over class labels with a categorical distribution: $p(y|\bpi) = Cat(y;\,\bpi)$. Its parameters, $\bpi$, follow a symmetric Dirichlet distribution, as we assume all classes to be equally probable: $p(\bpi) = Dir(\bpi;\, \alpha_{p}\!\cdot\!\bff{1}_{L})$. As the Dirichlet distribution is conjugate to the categorical, the posterior will also be a Dirichlet distribution: $ p(\bpi|y) = Dir(\bpi;\,\alpha_{p}\!\cdot\!\bff{1}_{L} + \bff{c}_{y})$. Note that $\bff{c}_{y}$ is a one-hot representation of the class label $y$.

We use an isotropic Gaussian distribution for $q(\bff{z}|\bff{x})$ and a Dirichlet for $q(\bpi|\bff{x})$. The neural networks that parametrise both of these approximate posteriors share parameters $\phi$. 
\begin{gather}
q_{\phi}(\bpi|\bff{x}) =  Dir(\bpi;\,\alpha_{q}\!\cdot\!\bff{\alpha(x; \phi)})\\ q_{\phi}(\bff{z}|\bff{x}) = \mathcal{N(\bff{z};\,\bff{\bmu(x; \phi)}, \bff{\bsigma(x; \phi)\!\cdot\!I})}
\end{gather}
We define $\bff{z}$ as the concatenation of two vectors: $\bff{z} = [\bff{z}_{c}^{\intercal}, \bff{z}_{s}^{\intercal}]^{\intercal}$. The class-dependent variables, $\bff{z}_{c}$, are the concatenation of a set of latent vectors, each of which is associated with a different class: $\bff{z}_{c} = [\bff{z}_{c1}^{\intercal}, \bff{z}_{c2}^{\intercal},\,... ,\,\bff{z}_{cL}^{\intercal}]^{\intercal}$. We want $\bff{z}_{c}$ to explain intraclass factors of variation. We refer to $\bff{z}_{s}$ as shared latent variables. They will express generative factors that are common to multiple classes. This allows for an efficient use of training samples, as these factors do not need to be learned separately for each input mode.

\begin{figure}[htb]
  \centering
  \resizebox{1\columnwidth}{!}{%
    \begin{tabular}{c@{\hskip 0.2in}c@{\hskip 0.2in}c}
      \begin{tikzpicture}
            
              \node[obs]                               (x1) {$\bff{x}$};
              \node[latent, below=of x1, xshift=-1.2cm] (pi) {$\bpi$};
              \node[latent, below=of x1, xshift=1.2cm]  (z1) {$\bff{z}$};
              \node[latent, below=of pi]  (y) {$y$};

              \edge {x1} {pi,z1} ; %
              \edge {pi} {y}
            
            \end{tikzpicture}  &
      \begin{tikzpicture}
            
              \node[obs]                               (x1) {$\bff{x}$};
              \node[latent, below=of x1, xshift=-1.2cm] (zc) {$\bff{z}_{cy}$};
              \node[latent, below=of x1, xshift=1.2cm]  (zs) {$\bff{z}_{s}$};
              \node[obs, below=of zc] (y) {$y$};
              \node[latent, below=of zs]  (z) {$\bff{z}$};

              \edge {zc, zs} {x1}; %
              \edge {y, z} {zc}; %
              \edge {z} {zs}; %
            
            
            \end{tikzpicture}   &
            \begin{tikzpicture}
            
              \node[obs]                               (x1) {$\bff{x}$};
              \node[latent, below=of x1, xshift=-1.2cm] (zc) {$\bff{z}_{cy}$};
              \node[latent, below=of x1, xshift=1.2cm]  (zs) {$\bff{z}_{s}$};
              \node[latent, below=of zc] (y) {$y$};
              \node[latent, below=of zs]  (z) {$\bff{z}$};

              \edge {zc, zs} {x1}; %
              \edge {y, z} {zc}; %
              \edge {z} {zs}; %
            
            
            \end{tikzpicture}\\[0.25cm]
        a) Encoder model.   &     b) Train-time decoder model.  &    c) Test-time decoder model.
    \end{tabular} 
    }
  \caption{\label{fig:graphical_model}Graphical model of N-VAE. The model is trained using the true class labels $y$. At test time, $y$ is inferred from $\bff{x}$.}
  \vskip -0.1in
\end{figure}
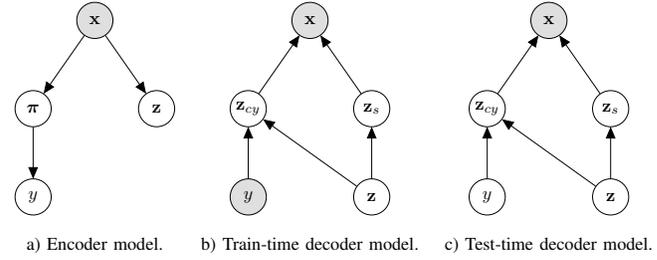

For a given class $y$, we compute the variable $\bff{z}_{cy} = vec(  \bff{c}_y \odot [\bff{1}_L, [\bff{z}_{c1}, \bff{z}_{c2},\,... ,\,\bff{z}_{cL}]^{\intercal}] )$.\footnote{We refer to the element-wise product with broadcasting by $\odot$. $vec(\cdot)$ denotes the vectorization of a matrix into a column vector.} The input, $\bff{x}$, is regenerated from $[\bff{z}_{cy}^{\intercal}, \bff{z}_{s}^{\intercal}]^{\intercal}$. The one-hot class label masks out the class-dependent latent variables for all non-relevant classes. It also introduces a class-dependent bias in the decoder's first activation layer when $\bff{c}_{y}$ is multiplied with its corresponding weight vector. The effects of this term are discussed in appendix B. The decoder is parametrized by $\theta$, taking the form $p_{\theta}(\bff{x}|y, \bff{z})\!=\!f(\bff{x}; y, \bff{z}, \bff{\theta})$. The complete graphical model is displayed in \cref{fig:graphical_model}. N-VAE is closely related to \cite{DIVA}, where domain dependent, class dependent and shared information are captured in separate latent spaces.

\begin{figure*}[ht]
\begin{center}
\centerline{\includegraphics[width=0.9\linewidth]{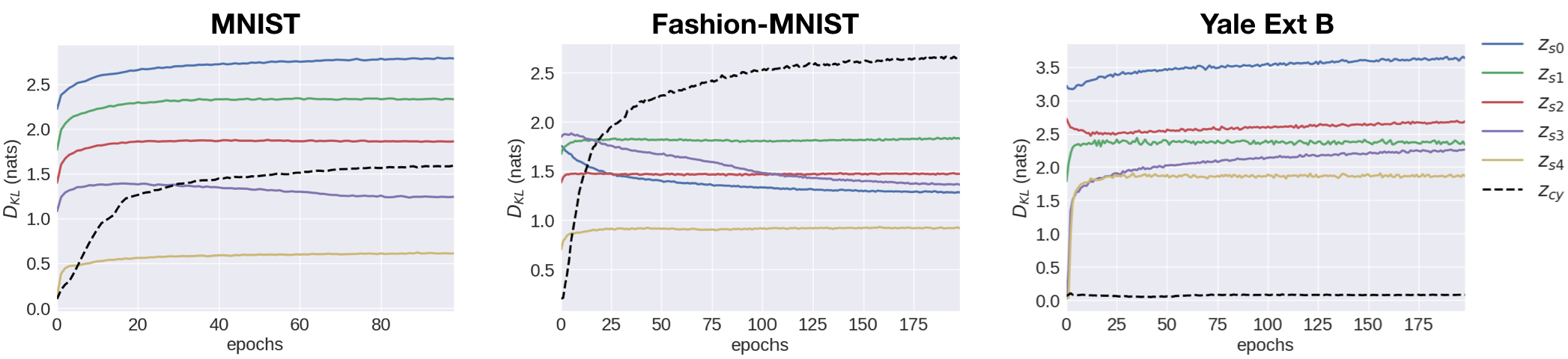}}
\caption{KL cost for each of N-VAE's latent dimensions for each of the datasets under consideration. $\bff{z}_{cy}$ refers to the KL cost of the class-dependent latent variable corresponding to the correct class.}
\label{fig:KL_per_dataset}
\end{center}
\vskip -0.2in
\end{figure*}

\subsection{Resulting objective function and optimization}

N-VAE's encoder is both a feature extractor and a classifier. For two Dirichlet distributions, $D_{KL}(q_{\phi}(\bpi|\bff{x})\,\|\, p(\bpi|y))$ can be upper bounded by the inferred class label's negative log likelihood: $-\log(q_{\phi}(y|\bff{x})) + k$, with $k$ being a constant. Using this approximation and dropping all constants with respect to model parameters, we optimize \cref{eq:ELBOdecomposed} as:
\begin{align}
\mathcal{L}_{obj}(\bff{x}, y) &=\EX_{q_{\phi}(\bff{z}|\bff{x})}[\log p_{\theta}(\bff{x}| y, \bff{z})]\notag\\
&- D_{KL}(q_{\phi}(\bff{z} |\bff{x})\,\|\,\ p(\bff{z})) + \log(q_{\phi}(y|\bff{x}))\label{eq:objective}
\end{align}
The KL divergence between the approximate posterior over $\bff{z}$ and its prior distribution is calculated for all dimensions of $\bff{z}\!=\![\bff{z}_{c}^{\intercal}, \bff{z}_{s}^{\intercal}]^{\intercal}$. However, not all dimensions of $\bff{z}_{c}$ are used to reconstruct every input. Most of them are masked out. The encoder learns to match the prior distribution for all dimensions of $\bff{z}_{c}$ which are not assigned to the correct class, making them non-informative. In other words, the encoder learns to model class-specific factors of variation.

\section{Experiments}\label{sec:results}
We consider three datasets which are separated into discrete classes: MNIST, Fashion-MNIST, and the cropped Extended Yale Face Dataset B. We preprocess the latter  by only keeping images with illumination angles smaller than 100\textdegree and scaling images to $64{\times}64$, \cite{yale_faces_ext_B}.

The training objective is reweighed in order to discourage $\bff{z}_{c}$ from learning factors of variation that are common to multiple classes:
\begin{align}
\mathcal{L}_{\beta_{c}} &=\EX_{q_{\phi}(\bff{z}|\bff{x})}[\log p_{\theta}(\bff{x}| y, \bff{z})]- D_{KL}(q_{\phi}(\bff{z}_{s} |\bff{x})\,\|\,\ p(\bff{z}))\notag\\ &- \beta_{c}D_{KL}(q_{\phi}(\bff{z}_{c} |\bff{x})\,\|\,\ p(\bff{z})) + \log(q_{\phi}(y|\bff{x}))\label{eq:beta_objective}
\end{align}
We set $\beta_{c}=2$. However, we find that N-VAE is relatively insensitive to the value of this parameter as long as $\beta_{c} > 1$.

\subsection{Detecting class-dependent factors of variation}\label{sec:detection}
In MNIST, there exist both class-exclusive and shared generative factors. However, in Fashion-MNIST, most of the continuous factors of variation seem to be intra-class. Classes such as sandals and t-shirts only share the factor of colour intensity. In Yale Ext B, the only continuous factors of variation are the elevation and azimuth of image illumination.

\begin{figure}[ht]
\vskip 0.05in
\begin{center}
\centerline{\includegraphics[width=1.05\columnwidth]{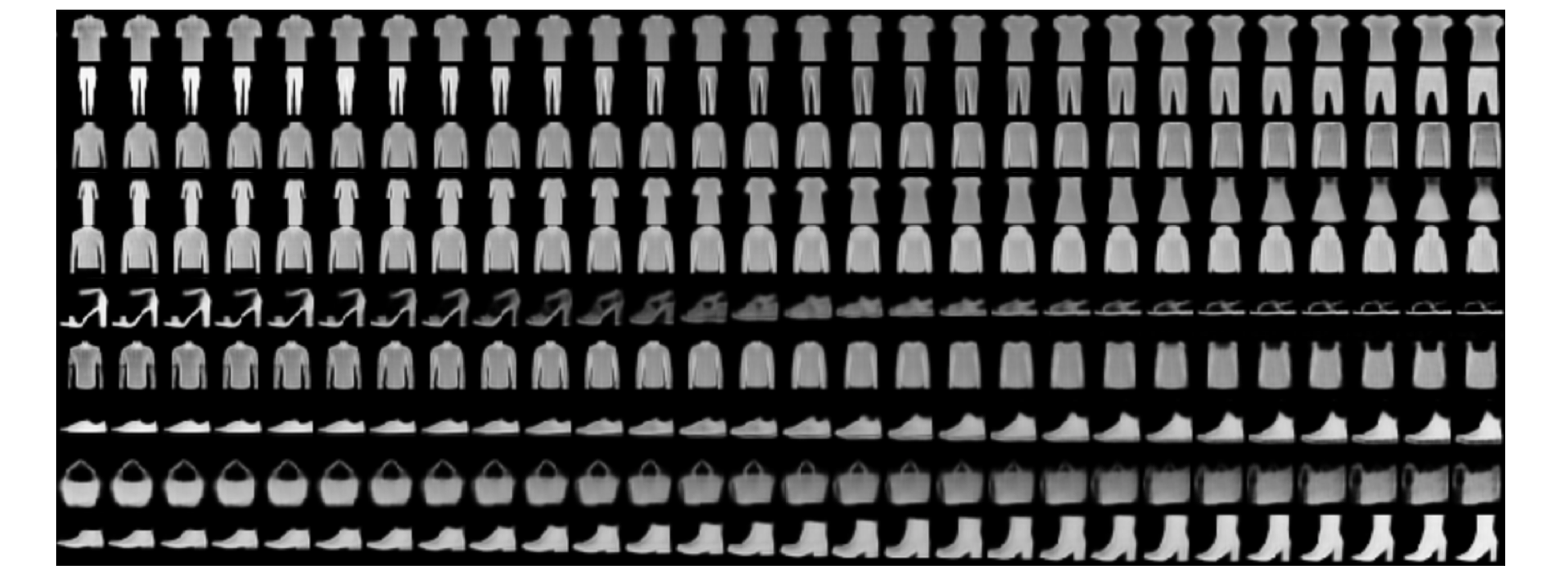}}
\caption{\label{fig:class_dependent_FMNIST}Fashion-MNIST class-dependent latent traversals produced by N-VAE by sampling $\bff{z}_{c}$ uniformly from $(-3, 3)$.}
\end{center}
\vskip -0.1in
\end{figure}

\begin{figure}[h]
\begin{center}
\centerline{\includegraphics[width=1\columnwidth]{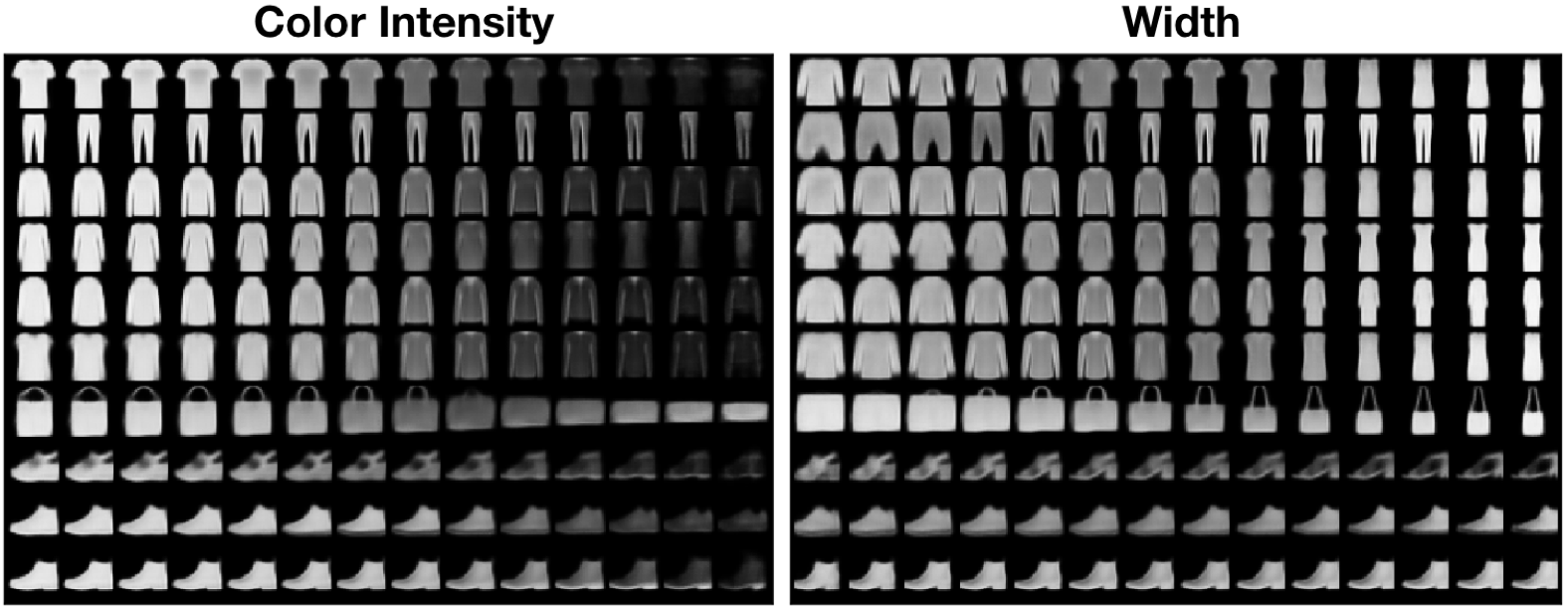}}
\caption{Shared latent traversals produced by N-VAE when trained on Fashion-MNIST.}
\label{fig:FMNIST_shared_latents}
\end{center}
\vskip -0.1in
\end{figure}

N-VAE's KL divergence values for $\bff{z}_{s}$ and $\bff{z}_{c}$ act as detectors for class-dependent and shared information, \cref{fig:KL_per_dataset}. The class-dependent variables carry the most information for Fashion-MNIST. For Yale Ext B, they are practically uninformative.

\subsection{Disentangling generative factors}

\begin{figure}[h]
\begin{center}
\centerline{\includegraphics[width=0.95\columnwidth]{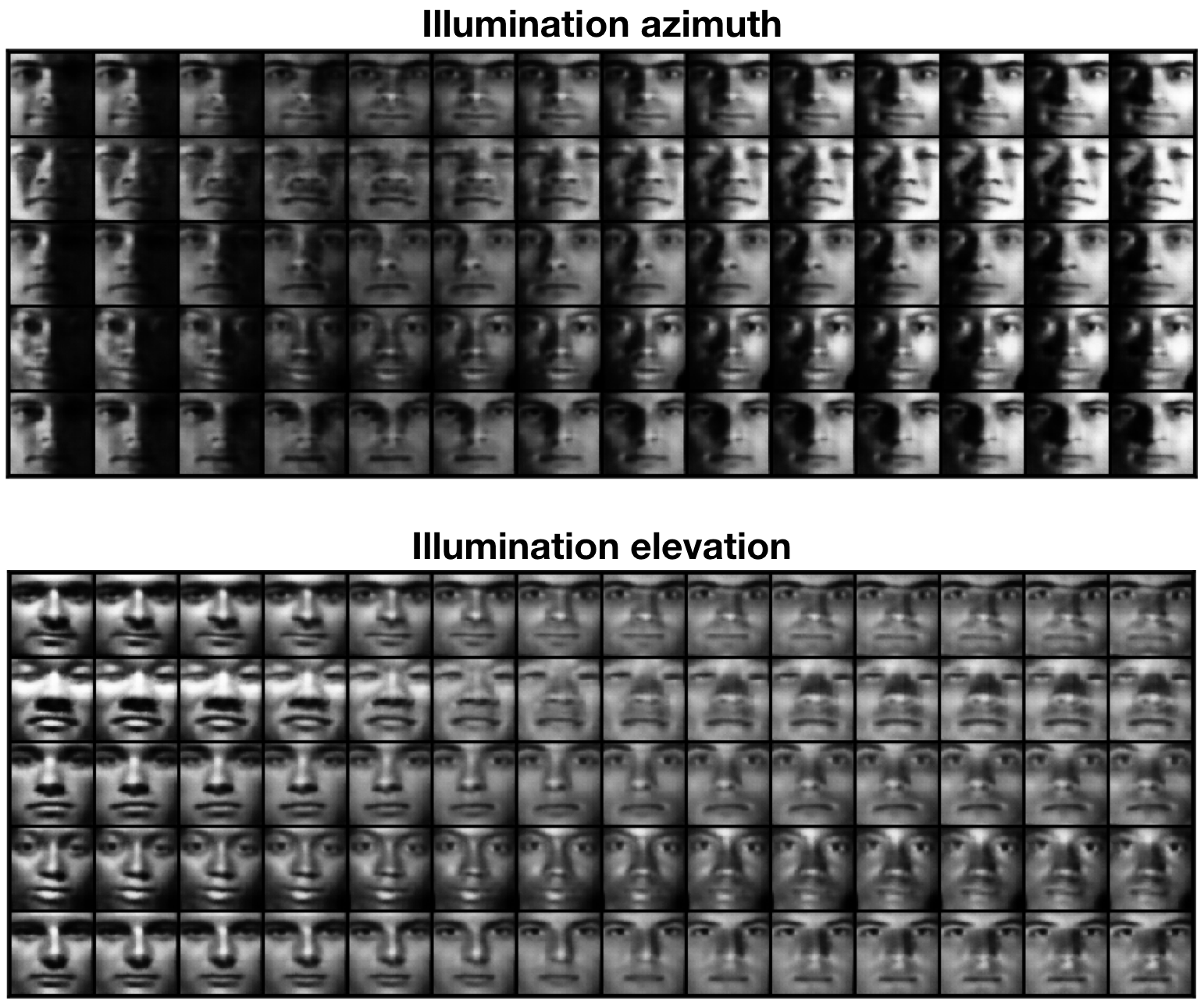}}
\caption{Shared latent traversals produced by N-VAE when trained on Yale Ext B. We show these for five classes out of 28.}
\label{fig:face_latents_traversal}
\end{center}
\vskip -0.1in
\end{figure}

Latent traversals for Fashion-MNIST's class-dependent variables are shown in \cref{fig:class_dependent_FMNIST}. In \cref{fig:FMNIST_shared_latents}, we show shared latent space traversals for the same dataset. The only generative factors which are shared among all classes seem to be color intensity and object width. In \cref{fig:face_latents_traversal}, we show shared latent space traversals for Yale Ext B. We obtain all latent traversals by sampling values of $\bff{z}_{s}$ uniformly from the range $(-3, 3)$.

We do not provide class-dependent latent space traversals for Yale Ext B, as $\bff{z}_{c}$ learns to be uninformative for this dataset. This is reflected in the values of $D_{KL}(q_{\phi}(\bff{z}_{c} |\bff{x})\,\|\,\ p(\bff{z}))$ from \cref{fig:KL_per_dataset}. Changing $\bff{z}_{c}$ does not affect reconstructions.

 By explicitly modelling multimodally distributed generative factors through class-dependent latent variables, the shared latent space is allowed to better match the factorial prior. Consequently, N-VAE is able to learn disentangled representations without elaborate modifications to the ELBO or extensive hyerparameter tuning.

\begin{figure*}[ht]
\begin{center}
\centerline{\includegraphics[width=1\linewidth]{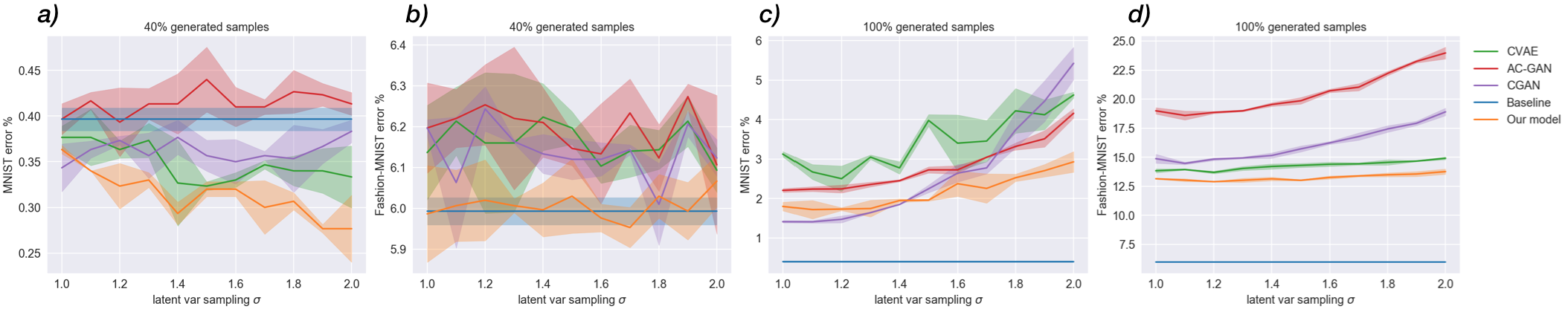}}
\caption{Mean and standard deviation of the classification error obtained when training with data from each generative model. All experiments have been repeated three times. We show results for probabilities of substituting original datapoints with artificial ones of $0.4$ (a value we found to perform well) and $1$. We vary $\sigma$ from 1 to 2. 'Baseline' refers to the classifier trained without artificial data.}
\label{fig:gen_results}
\end{center}
\vskip -0.2in
\end{figure*}
\subsection{Augmenting Datasets with Generated Samples}\label{sec:generation_experiment}

We train classifiers with different proportions of original and artificial (generated) samples and evaluate them using the original test sets. With this method, we not only test the discriminability of our generations but also how much of the original datasets' variability they capture. A more diverse set of generated images will better constrain the classifiers' parameters, resulting in higher accuracy. 

We compare N-VAE with the conditional VAE (CVAE), \cite{CVAE}, the conditional GAN (CGAN), \cite{CGAN} and the auxiliary classifier GAN (AC-GAN), \cite{AC-GAN}. All of these models use class labels during training and allow for class-conditional sample generation. Latent variables are sampled from $\mathcal{N(\bff{z};\,\bff{0}, \sigma\!\cdot\!\bff{I})}$. A larger $\sigma$ allows sampling from regions of latent space which are less likely under the trainset. A large enough value should allow for the extrapolation of learned generative factors, producing more extreme characteristics or new combinations of features. Our intention is to show that generative models generalize better when an appropriate latent space structure is used, not to propose a new data augmentation scheme.

On MNIST, including artificial samples has a regularization effect for most generative models as is shown in \cref{fig:gen_results} a); Seeing more plausible data makes it more difficult for the classifiers to overfit to specific training samples. However, N-VAE is the only model for which there is a consistent decrease in classification error as $\sigma$ increases. It is able to produce feature configurations that complement those of the original training set. Fashion-MNIST is a more complex dataset which is harder to explain for generative models. In plot b), we see that using samples generated with N-VAE decreases the baseline classifier's performance the least on this dataset.

We observe a significant performance decrease when training classifiers exclusively on artificial data in plots c) and d) from \cref{fig:gen_results}. On MNIST, the conditional GAN produces the most useful samples for $\sigma=1$, N-VAE is a close second. On Fashion-MNIST, N-VAE provides the best results. In all scenarios, N-VAE performs best as $\sigma$ increases.

\section{Conclusion}
N-VAE explains data in terms of shared latent variables and class-dependent latent variables. We have shown the utility of this type of model for identifying and disentangling class-dependent generative factors in data. In our data augmentation experiment, N-VAE is the only model that provides more informative samples for classification as $\sigma$ increases. This suggests that N-VAE is able to produce samples with new combinations of features, showing its capacity for generalization.

\section*{Acknowledgment}
This work has been supported by the Spanish Ministry of Economy and Competitiveness and the European Social Fund through the project TIN2017-85854-C4-1-R, by the Government of Aragon (Reference Group T36\_17R) and co-financed with Feder 2014-2020 "Building Europe from Aragon."

\bibliographystyle{IEEEtran}  
\bibliography{example_paper}

\newpage

\appendix

\section*{A: Derivation of \cref{eq:jointELBO} and \cref{eq:ELBOdecomposed}}\label{app:derive_our_objective}
First, we show that our objective is a lower bound on the joint log-likelihood $\log p(\bff{x}, y)$. Our starting point is the KL divergence between the true and approximate posteriors over $\bff{z}$ and $\bpi$, as this is the quantity which we wish to minimise. 
\begin{align*}
         &D_{KL}( q(\bff{z}, \bpi|\bff{x}, y)\,\|\,p(\bff{z}, \bpi|\bff{x}, y) )\notag\\ &= \EX_{q(\bff{z}, \bpi|\bff{x}, y)} [\log q(\bff{z}, \bpi|\bff{x}, y) - \log p(\bff{z}, \bpi|\bff{x}, y)]\notag\\
        &= \EX_{q(\bff{z}, \bpi|\bff{x}, y)} [\log q(\bff{z}, \bpi|\bff{x}, y)\notag\\ &- \log p(\bff{x}, y, \bff{z}, \bpi)] + \log p(\bff{x}, y)\notag\\
        &= -\mathcal{L}(\bff{x}, y) + \log p(\bff{x}, y)
\end{align*}
Using the non-negativity of the KL divergence, we can write: $\mathcal{L}(\bff{x}, y) \leq \log p(\bff{x}, y)$.

We make the following modelling choices: 
\begin{itemize}
    \item $\bff{x}$ is conditionally independent of $\bpi$ given $y$; $p(\bff{x} | \bff{z}, y, \bpi) = p(\bff{x} | \bff{z}, y)$.
    \item Our inference model $q$ only takes $\bff{x}$ as an input. Therefore, $q(\bpi | \bff{x}, y) = q(\bff{z} | \bff{x})$ and $q(\bpi | \bff{x}, y) = q(\bpi | \bff{x})$.
\end{itemize}
We can rewrite \cref{eq:jointELBO} in the following manner:
\begin{align*}
    &\log p(\bff{x}, y) \geq \mathcal{L}(\bff{x}, y)\notag\\ &=\EX_{q(\bff{z}, \bpi|\bff{x}, y)} [-\log q(\bff{z}, \bpi|\bff{x}, y) + \log p(\bff{x}, y, \bff{z}, \bpi)]\notag\\
    &= \EX_{q(\bff{z}, \bpi|\bff{x}, y)} [-\log q(\bff{z}, \bpi|\bff{x}, y)\notag\\
    &+ \log(p(\bff{x}| y, \bpi, \bff{z}) p(\bff{z}) p(\bpi|y) p(y))]\notag\\
    &= \EX_{q(\bff{z}, \bpi|\bff{x}, y)}[\log p(\bff{x}| y, \bpi, \bff{z})]\notag\\
    &- \EX_{q(\bff{z}, \bpi|\bff{x}, y)}[\log q(\bff{z} |\bff{x}, y) - \log p(\bff{z})]\notag\\ &- \EX_{q(\bff{z}, \bpi|\bff{x}, y)}[\log q(\bpi|\bff{x}, y) - \log p(\bpi|y)]\notag\\
    &+ \EX_{q(\bff{z}, \bpi|\bff{x}, y)}[\log p(y)]\notag\\
    &= \EX_{q(\bff{z}|\bff{x})}[\log p(\bff{x}| y, \bff{z})] - D_{KL}(q(\bff{z} |\bff{x})\,\|\,\ p(\bff{z}))\notag\\
    &- D_{KL}(q(\bpi|\bff{x})\,\|\, p(\bpi|y)) + \log p(y)
\end{align*}

\vspace{0.1in}
\section*{B: The bias term in the decoder's input}\label{app:bias}

$\bff{z}_{cy}$ is obtained as $vec(  \bff{c}_y \odot [\bff{1}_L, [\bff{z}_{c1}, \bff{z}_{c2},\,... ,\,\bff{z}_{cL}]^{\intercal}] )$. In this expression, we can see that $\bff{z}_{cy}$ is composed of the latent vectors predicted by the encoder, of which all but one is masked out, and a one-hot vector which results from $\bff{c}_y \odot \bff{1}_L=\bff{c}_y$. The latter explicitly informs the decoder about the class identity of the sample to be reconstructed. When multiplied with the corresponding weight layer, the one-hot term adds a class-dependent bias term $b_{y}$ to the decoder's first activation layer. 

\begin{figure}[h]
\begin{center}
\centerline{\includegraphics[width=0.8\columnwidth]{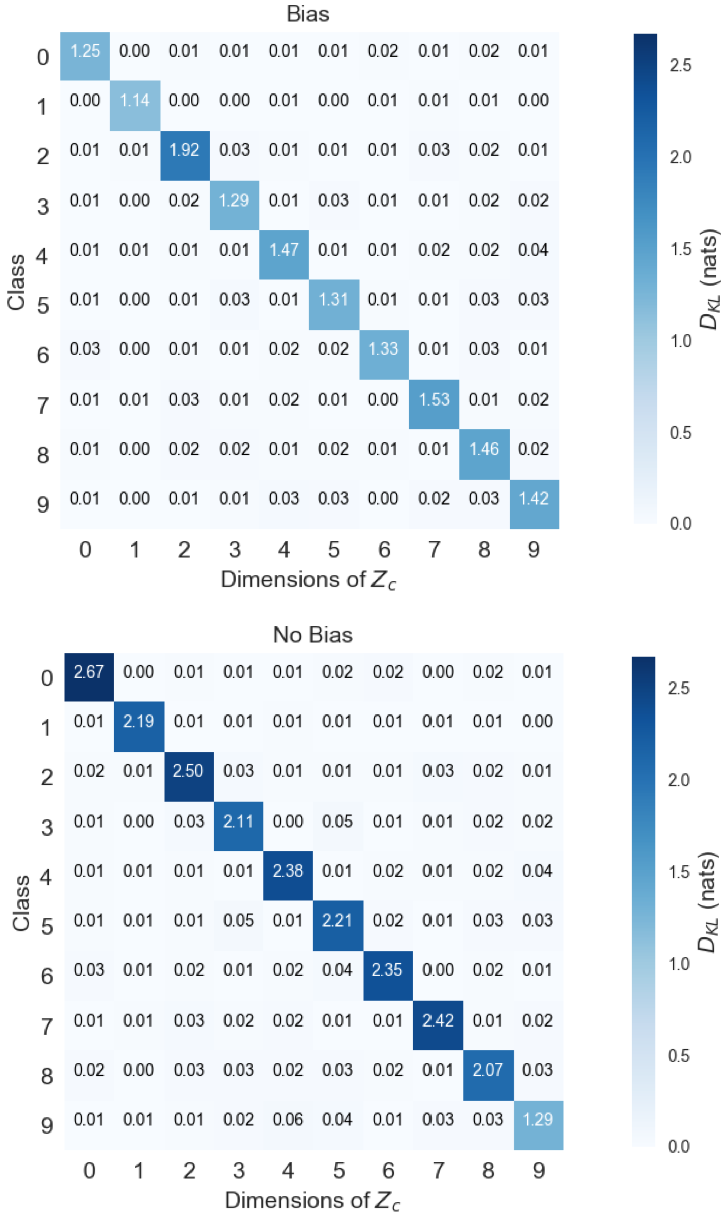}}
\caption{Mean value of each dimension of $D_{KL}(q_{\phi}(\bff{z}_{ci} |\bff{x})\,\|\,\ p(\bff{z}))$ for each class of the MNIST test set. The encoder learns to classify samples implicitly by only making the correct dimensions of $\bff{z}_{c}$ informative. The addition of a class-dependent bias term to the decoder allows for the distribution of the class-dependent latent variables to be closer to the Gaussian prior.}
\label{fig:bias_confusion}
\end{center}
\vskip -0.1in
\end{figure}

The most likely value under the prior $p(\bff{z}_{c})= \mathcal{N}(\bff{z}_{c};\,\bff{0}, \bff{I})$ is $0$. However, when we mask out latent variables, we set most of them to this value. This makes masked latent variables indistinguishable from samples from $p(\bff{z}_{c})$. Without $b_{y}$, the encoder is forced to predict distributions which place a lot of mass on large values of $\bff{z_{c}}$ in order to communicate the class of $\bff{x}$ to the decoder. In \cref{fig:bias_confusion}, we show how the KL divergence grows for class-dependent latent variables when we remove $\bff{c}_{y} \odot \bff{1}_{L}$ and thus, the bias, from $\bff{z}_{cy}$. The explicit encoding of discrete information allows us to describe the data's class-dependent generative factors with distributions that more closely resemble the prior. It allows for smooth interpolation of representations in the class-conditioned latent space around zero.

\section*{C: N-VAE samples with $\sigma = 1.4$}\label{app:high_var_generations}


\begin{figure}[ht]
\begin{center}
\centerline{\includegraphics[width=0.55\columnwidth]{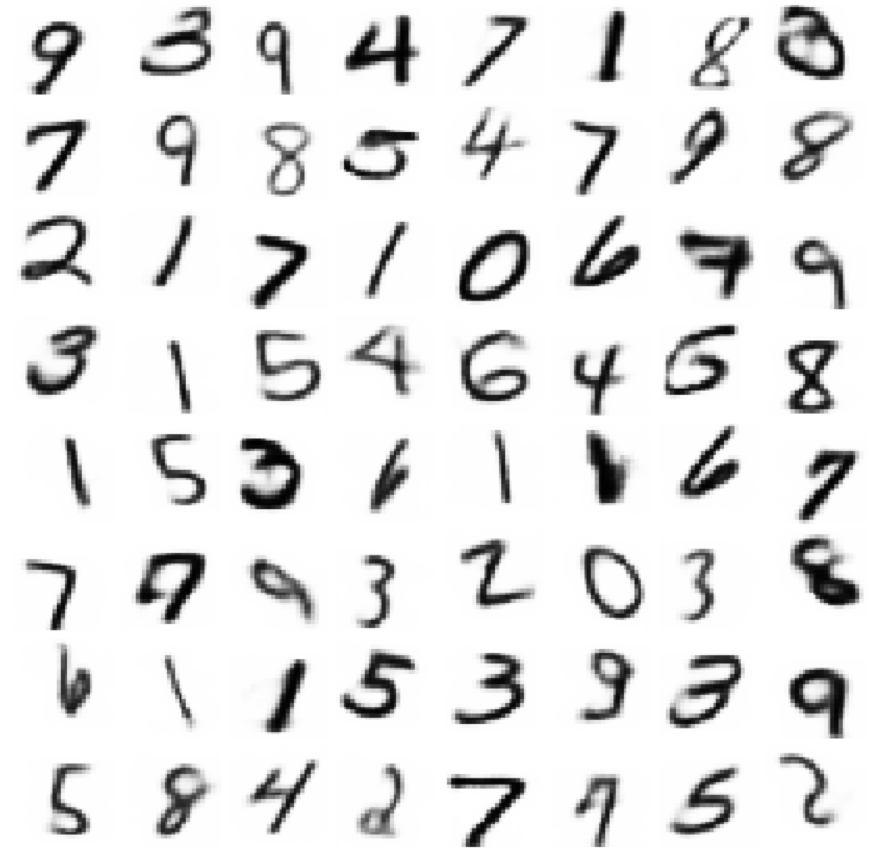}}
\caption{Digits generated with N-VAE by sampling latent variables from  $\mathcal{N}(\bff{z};\,\bff{0}, 1.4\!\cdot\!\bff{I})$.}
\label{fig:high_var_gens}
\end{center}
\vskip -0.2in
\end{figure}

\end{document}